\documentclass[10pt,twocolumn,letterpaper]{article}

\usepackage{cvpr}
\usepackage{times}
\usepackage{epsfig}
\usepackage{graphicx}
\usepackage{amsmath}
\usepackage{amssymb}
\usepackage{tikz, tikzsymbols}
\usetikzlibrary{backgrounds, positioning, arrows.meta, arrows, calc, fit, decorations.pathreplacing}
\usepackage{booktabs}
\usepackage{multirow}
\usepackage{tabularx}
\usepackage{makecell}
\usepackage[breaklinks=true,bookmarks=false]{hyperref}

\newcommand{\etwo}{E_\textit{2D}}
\newcommand{\ethree}{E_\textit{3D}}
\newcommand{\dtwo}{D_\textit{2D}}
\newcommand{\dthree}{D_\textit{3D}}
\newcommand{\epose}{E_\textit{pose}}

\newcommand{\R}{\mathbb{R}}
\newcommand{\loss}[1]{\mathcal{L}_\textit{#1}}

\newcolumntype{C}{>{\centering\arraybackslash}X}
\newcolumntype{L}{>{\raggedright\arraybackslash}X}

\newlength{\hf}
\newlength{\hb}
\newlength{\h}
\newlength{\w}
\newlength{\de}
\definecolor{cencdec}{HTML}{64b0b0}
\definecolor{cwarp}{HTML}{5656a3}

\newcommand{\BlockTwoToThree}[5]{%
  \begingroup%
  \setlength{\hf}{#1}%
  \setlength{\hb}{#2}%
  \setlength{\w}{#3}%
  \setlength{\de}{#4}%
  \def\text{#5}%
  \begin{tikzpicture}
    \draw[fill=cencdec!20] (0,0.5\hf) -- (\w,0.5\hb) -- (\w,-0.5\hb) -- (0,-0.5\hf) -- cycle;
    \draw[overlay,fill=cencdec!20] (\w,-0.5\hb) -- (\w+\de,-0.5\hb+\de) -- (\w+\de,0.5\hb+\de) -- (\w,0.5\hb);
    \draw[overlay,fill=cencdec!20] (\w+\de,0.5\hb+\de) -- (0,0.5\hf) -- (\w,0.5\hb);
    \node[align=center,anchor=center,fill=cencdec!20] at (0.5\w,0) {\text};
  \end{tikzpicture}%
  \endgroup%
}
\newcommand{\BlockThreeToTwo}[5]{%
  \begingroup%
  \setlength{\hf}{#1}%
  \setlength{\hb}{#2}%
  \setlength{\w}{#3}%
  \setlength{\de}{#4}%
  \def\text{#5}%
  \begin{tikzpicture}
    \draw[fill=cencdec!20] (0,0.5\hf) -- (\w,0.5\hb) -- (\w,-0.5\hb) -- (0,-0.5\hf) -- cycle;
    \draw[overlay,fill=cencdec!20](0,0.5\hf) -- (0+\de,0.5\hf+\de) -- (\w,0.5\hb) -- (\w,0.5\hb);
    \node[align=center,anchor=center] at (0.5\w,0) {\text};
  \end{tikzpicture}%
  \endgroup%
}
\newcommand{\BlockTwoToTwo}[5]{%
  \begingroup%
  \setlength{\hf}{#1}%
  \setlength{\hb}{#2}%
  \setlength{\w}{#3}%
  \setlength{\de}{#4}%
  \def\text{#5}%
  \begin{tikzpicture}
    \draw (0,0.5\hf) -- (\w,0.5\hb) -- (\w,-0.5\hb) -- (0,-0.5\hf) -- cycle;
    \node[align=center,anchor=center] at (0.5\w,0) {\text};
  \end{tikzpicture}%
  \endgroup%
}
\newcommand{\BlockThreeToThree}[4]{%
  \begingroup%
  \setlength{\h}{#1}%
  \setlength{\w}{#2}%
  \setlength{\de}{#3}%
  \def\text{#4}%
  \begin{tikzpicture}
    \draw (0,0.5\h) -- (\w,0.5\h) -- (\w,-0.5\h) -- (0,-0.5\h) -- cycle;
    \draw[overlay] (0,0.5\h) -- (0+\de,0.5\h+\de) -- (\w+\de,0.5\h+\de) -- (\w+\de,-0.5\h+\de) -- (\w,-0.5\h);
    \draw[overlay] (\w,0.5\h) -- (\w+\de,0.5\h+\de);
    \node[align=center,anchor=center] at (0.5\w,0) {\text};
  \end{tikzpicture}%
  \endgroup%
}
\newcommand{\BlockThree}[4]{%
  \begingroup%
  \setlength{\h}{#1}%
  \setlength{\w}{#2}%
  \setlength{\de}{#3}%
  \def\img{#4}%
  \begin{tikzpicture}
    \draw[overlay] (0,0.5\h) -- (0+\de,0.5\h+\de) -- (\w+\de,0.5\h+\de) -- (\w+\de,-0.5\h+\de) -- (\w,-0.5\h);
    \draw (0,0.5\h) -- (\w,0.5\h) -- (\w,-0.5\h) -- (0,-0.5\h) -- cycle;
    \draw[overlay] (\w,0.5\h) -- (\w+\de,0.5\h+\de);
    \node[overlay,align=center,anchor=center] at (0.5\w+0.5\de,0.5\de) {\img};
  \end{tikzpicture}%
  \endgroup%
}
\newcommand{\BlockThreeColored}[4]{%
	\begingroup%
	\setlength{\h}{#1}%
	\setlength{\w}{#2}%
	\setlength{\de}{#3}%
	\def\img{#4}%
	\begin{tikzpicture}

	\draw[overlay,fill=cwarp!20] (0,0.5\h) -- (0+\de,0.5\h+\de) -- (\w+\de,0.5\h+\de) -- (\w+\de,-0.5\h+\de) -- (\w,-0.5\h);
	\draw[fill=cwarp!20] (0,0.5\h) -- (\w,0.5\h) -- (\w,-0.5\h) -- (0,-0.5\h) -- cycle;
	\draw[overlay] (\w,0.5\h) -- (\w+\de,0.5\h+\de);
	\node[overlay,align=center,anchor=center] at (0.5\w+0.5\de,0.5\de) {\img};
	\end{tikzpicture}%
	\endgroup%
}
\newcommand{\BlockTwo}[3]{%
  \begingroup%
  \setlength{\h}{#1}%
  \setlength{\w}{#2}%
  \def\img{#3}%
  \begin{tikzpicture}
    \node[overlay,align=center,anchor=center] at (0.5\w,0) {\img};
    \draw (0,0.5\h) -- (\w,0.5\h) -- (\w,-0.5\h) -- (0,-0.5\h) -- cycle;
  \end{tikzpicture}%
  \endgroup%
}

\newcommand{\PAR}[1]{\vskip1pt \noindent {\bf #1~}}

\cvprfinalcopy

\def\cvprPaperID{9}

\ifcvprfinal\fi
\begin{document}

\title{Reposing Humans by Warping 3D Features}

\author{Markus Knoche \qquad Istv\'an S\'ar\'andi \qquad Bastian Leibe\\
RWTH Aachen University, Germany\\
{\tt\small \{knoche,sarandi,leibe\}@vision.rwth-aachen.de}
}

\maketitle
\thispagestyle{empty}

\begin{abstract}
We address the problem of reposing an image of a human into any desired novel pose.
This conditional image-generation task requires reasoning about the 3D structure of the human, including self-occluded body parts.
Most prior works are either based on 2D representations or require fitting and manipulating an explicit 3D body mesh.
Based on the recent success in deep learning-based volumetric representations, we propose to implicitly learn a dense feature volume from human images, which lends itself to simple and intuitive manipulation through explicit geometric warping.
Once the latent feature volume is warped according to the desired pose change, the volume is mapped back to RGB space by a convolutional decoder.

Our state-of-the-art results on the DeepFashion and the iPER benchmarks indicate that dense volumetric human representations are worth investigating in more detail.
\end{abstract}

\section{Introduction}

The ability to freely change a human's pose in an image opens the door to a variety of applications, from generating large crowds or performing stunts in filmmaking to data augmentation for human-centric computer vision tasks.
State-of-the-art approaches to this problem employ fully-convolutional neural networks.
However, convolutional features tend to strongly depend on the input pixels near the center of the receptive field and CNNs often fail to move information over large distances.
This makes person reposing difficult when input and target pose differ strongly, as the appearance information of the various body parts needs to move to different places compared to their position in the input image.
To tackle this, many recent approaches apply some form of explicit transformations.
Some warp 2D features such that they become aligned to the target pose, which is also specified in 2D \cite{Balakrishnan18CVPR,Siarohin18CVPR,Horiuchi19MVA,Dong18NIPS,Neverova18ECCV,Grigorev19CVPR}.
We argue that this is insufficient to capture 3D human pose changes.

Mesh-based approaches fit a 3D body model to the input, infer the texture and render the mesh in the target pose \cite{Zanfir18CVPR,Liu19ICCV}.
While capturing the 3D aspect, this has the downside that a specific human might not be captured well by a general model, for example due to uncommon hairstyles and spacious clothing.

Inspired by recent volumetric approaches for related tasks~\cite{Pavlakos17CVPR,Nguyen19ICCVW}, we propose a novel reposing method, illustrated in Fig.~\ref{fig:over}, which warps 3D volumetric CNN-features without requiring an explicit mesh model.
Using only a 2D image as input, our model implicitly learns a latent volumetric representation of the input person.
This representation is then warped using 3D transformations based on input and target pose to align it to the target pose.
We process the warped features along with 3D target pose heatmaps with a decoder, to synthesize the reposed image.

By ablation, we show the benefits of the two 3D aspects of our work: first the 3D warping, and second, representing the target pose in 3D.
Overall, our method achieves state-of-the-art scores on the commonly used DeepFashion and the newer iPER benchmarks.

\begin{figure}
\centering
\begin{tikzpicture}[node distance=.17cm,every node/.style={font=\sffamily\footnotesize}]
\definecolor{cwarp}{HTML}{5656a3}
  \def\imgs{1.1cm}
  \def\vols{.55cm}
  \def\widt{2cm}
  \def\dept{.15cm}
  \node[inner sep=0pt,fill=gray!7] (i) {\BlockTwo{\imgs}{\imgs}{\Strichmaxerl[4.5][-40][40][10][-50]}};
  \node[inner sep=0pt,right=of i] (e) {\BlockTwoToThree{\imgs}{\vols}{\widt}{\dept}{Lifting\\Encoder}};
  \node[inner sep=0pt,right=of e] (v1) {\BlockThreeColored{\vols}{\vols}{\dept}{\Strichmaxerl[3][-40][40][10][-50]}};
  \node[inner sep=0pt,right=of v1] (v2) {\BlockThreeColored{\vols}{\vols}{\dept}{\Strichmaxerl[3][-10][-50][60][-60]}};
  \node[inner sep=0pt,right=of v2] (d) {\BlockThreeToTwo{\vols}{\imgs}{\widt}{\dept}{Projection\\Decoder}};
  \node[inner sep=0pt, right=of d,fill=gray!7] (o) {\BlockTwo{\imgs}{\imgs}{\Strichmaxerl[4.5][-10][-50][60][-60]}};
  \draw[->] ($(v1.center)+(0,-1ex)$) to [bend right=60] ($(v2.center)+(2ex,-1.2ex)$);
  \draw[->] ($(v1.center)+(1.2ex,-1.4ex)$) to [bend right=40] ($(v2.center)+(0,-1.2ex)$);
  \draw[->] ($(v1.center)+(-.0ex,1.7ex)$) to [bend left=60] ($(v2.center)+(1.7ex,.1ex)$);
  \draw[->] ($(v1.center)+(2.1ex,1.4ex)$) to [bend left=40] ($(v2.center)+(0,.6ex)$);
  \node[align=center,anchor=north,inner sep=2pt] (pi) at ($(v1)!0.5!(v2)+(1ex,10ex)$) {Warp according to \\desired pose change};
  \draw[->] (pi) -- +(0,-4.5ex);
\end{tikzpicture}
\caption{Our encoder network implicitly learns a volumetric representation of the input person, such that 3D feature warping can be applied in the middle of the architecture to achieve reposing.}
\label{fig:over}
\end{figure}
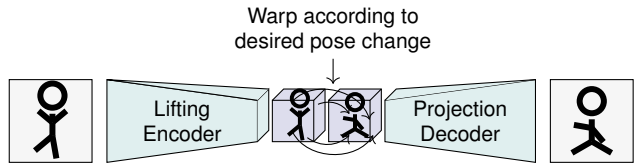
\begin{figure*}
	\centering
	\input{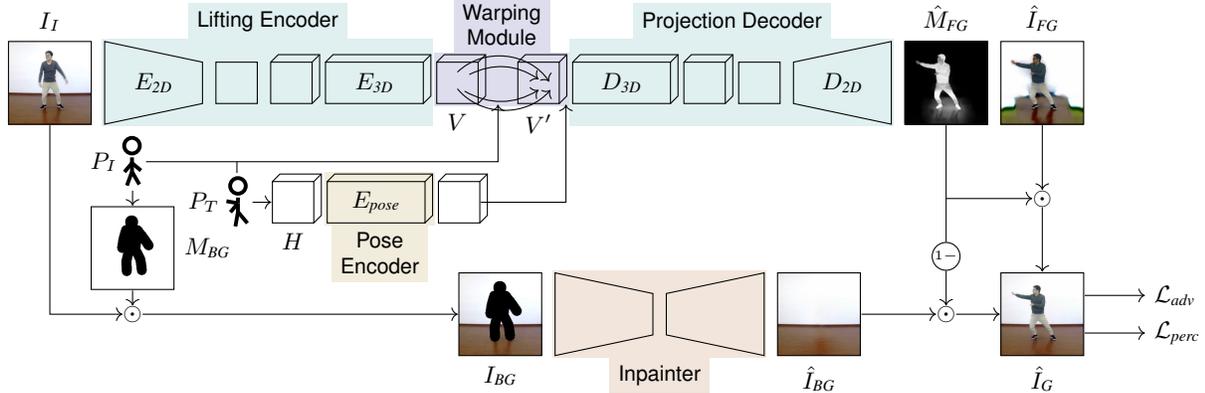}
	\caption{
		Generator architecture.
		The foreground stream learns 3D features from a 2D image and applies 3D feature warping.
		The result is combined with the target pose and projected to an RGBA image.
		Alpha blending with an inpainted background yields the final output.}
	\label{fig:architecture}
\end{figure*}

\section{Related Work}

Image generation methods have come a long way since the introduction of generative adversarial networks (GAN)~\cite{Goodfellow14NIPS}.
Building on Isola \etal's image-conditioned GAN~\cite{Isola17CVPR}, Ma \etal were first to tackle pose-conditioned person image generation~\cite{Ma17NIPS}.
They feed the image and 2D target pose heatmap through two stages: the first is trained with a pixel-wise $L_1$ loss, the second with an adversarial loss.
Lakhal \etal~\cite{Lakhal18ECCV} use two encoders in both stages, distinguishing between aligned and misaligned input in Stage I and between pose and images in Stage II.
Similar subdivisions are used in other works~\cite{Si18CVPR,Zhu19CVPR}.

The misalignment of input and target is tackled by explicit warping in many works.
Siarohin \etal~\cite{Siarohin18CVPR} use affine warps on the skip connections of a U-net architecture~\cite{Ronneberger15MICCAI}.
They mask out features corresponding to bodyparts based on the input pose and warp them to align with the target pose.
An extension is proposed in~\cite{Horiuchi19MVA}, adding self-attention layers, spectral normalization and a relativistic discriminator to the architecture.
In \cite{Balakrishnan18CVPR}, a similar transformation is applied directly on the input image, using learned soft masks.

Some methods warp based on body part segmentation or dense pose~\cite{Alp18CVPR} to encode input and target pose instead of using keypoint representations~\cite{Dong18NIPS,Neverova18ECCV,Grigorev19CVPR}.
This tells the network the exact shape of the target person, making the task simpler, but a dense target pose is not available in general.

The 3D mesh-based approach of~\cite{Zanfir18CVPR} fits a body model to the given person, back-projects pixels onto the mesh and transforms the mesh to the target pose.
Unseen texture is inpainted with a neural net.
Other methods~\cite{Liu19ICCV, Li2019CVPR} use meshes to compute a transformation flow from input to target pose, which is used to transform network features in 2D.

A line of works in 3D human pose estimation~\cite{Pavlakos17CVPR,Sun18ECCV,Luvizon18CVPR,Sarandi20FG} has shown that it is feasible to predict depth-related information from images in a volumetric representation (in that case volumetric body joint heatmaps), by a tensor reshaping operation.
We take this as inspiration to predict volumetric feature maps of humans in our work.

Human reposing can be viewed as a generalization of novel-view synthesis (NVS) from rigid to articulated pose.
As volumetric prediction has also been successfully applied for NVS~\cite{Sitzmann19CVPR,Nguyen19ICCVW}, we take this as further motivation to investigate the usefulness of a similar representation in reposing.

In contrast to the volumetric approach, a sparse 3D representation is used in \cite{Rhodin18ECCV} to learn NVS.
The encoder outputs an appearance feature vector and a 3D point cloud representing the pose.
After rotating the point cloud, the decoder transforms both back to an image from another view.
The implicitly learned point cloud is given to a shallow human pose estimation network, thereby reducing the amount of labeled pose estimation data needed.
Similar to our method, an implicitly learned 3D structure is explicitly transformed, however, instead of a point cloud which only represents the pose we transform volumetric features which also contain appearance information.

\section{Method}

Given an input image $I_I$ of a person and a target pose $P_T$, we aim at generating an image $\hat I_G$ of the person in pose $P_T$.
We use a two-stream generator network to tackle this problem, where the first stream reposes the person using our novel volumetric feature warping approach, while the second inpaints missing parts of the background.
To utilize the volumetric warping, the model has to estimate the depth of different bodyparts such that it can lift the corresponding features accordingly to a 3D volume.
This is learned implicitly from the 3D warping, we neither give depth information about the input pose to our model nor do we apply any explicit supervision with respect to the input pose.

\subsection{Architecture}

Our architecture consists of a lifting encoder, a 3D warping module, a projection decoder and a background inpainter as shown in Fig.~\ref{fig:architecture}.

The \textbf{lifting encoder} maps a 2D input image to 3D volumetric features.
The 2D input image $I_I$ is passed to a convolutional network $\etwo{}$ which outputs 2D feature maps $\etwo{}(I_I)\in \R^{H\times W \times D \cdot C}$.
A reshape operation splits the channel dimension of the resulting tensor into different depth layers, yielding the feature volume $F\in \R^{H\times W \times D \times C}$. This is similar to how joint heatmaps are estimated in~\cite{Pavlakos17CVPR}, but instead of heatmaps, we produce a latent feature volume.
$\etwo{}$ thus learns that different features in its output belong to different depths.
To further process these volumetric features, a 3D convolutional network ($\ethree{}$) is applied to yield $V\in \R^{H\times W \times D \times C}$.

The key element of our approach is our novel \textbf{3D warping module}, whose purpose is to shuttle voxel features to their target location.
It gets a feature volume $V \in \R^{H\times W \times D \times C}$, together with the 3D input and target pose $P_I, P_T\in \R^{J\times 3}$ which are given as 3D joint coordinates.
The input pose $P_I$ is used to create ten masks $M_i\in \{0,1\}^{H\times W \times D}$, one per bodypart.
Masks are generated by drawing capsular shapes between the joints corresponding to that bodypart, \eg, the lower left leg's mask is based on the left ankle and the left knee joints and the mask of the torso depends on the hips and shoulders.
We then create ten copies $V_i$ of the feature volume and apply the corresponding mask by voxel-wise multiplication, giving ten volumes, one per body part.
Next, we fit a transformation $T_i$ for each body part based on input and target joints.
We assume that each part moves rigidly, but as the scale of the person in pixel space may change, we also add a scale parameter.
The result is a 7-parameter Helmert transformation, estimated by least squares.
When a body part has only two joints, as for leg and arm parts, we use a third joint to specify the rotation around the body part's axis.
For example, the left lower arm's motion would only depend on left wrist and left elbow, so we use the left shoulder's position as an anchor.
The masked bodypart features are then warped according to the respective transformation using trilinear interpolation and combined using the maximum activation.
Given $M_i$ and $T_i$, the output feature volume of the warping module is
\[V'=\max_i T_i(M_i\odot V).\]
The \textbf{target pose encoder} $\epose{}$ feeds the target pose into our model.
Its input are Gaussian volumetric heatmaps $H\in\R^{H\times W\times D\times J}$, one per body joint.
The result is concatenated to the warped volumetric features and processed by the projection decoder.

Mirroring the lifting encoder, our \textbf{projection decoder} contains two parts $\dthree{}$ and $\dtwo{}$.
The 3D convolutional network $\dthree{}$ allows to enhance the warped features and also combines them with the output of the target pose encoder.
This volume is reshaped to 2D by combining depth and volumetric channels into a single channel dimension.
We then apply the second decoder network $\dtwo{}$, yielding the generated RGB image $\hat I_\textit{FG}$ together with a soft mask $\hat M_\textit{FG}$.

We apply a \textbf{background inpainter} stream, since our warping module masks bodyparts and only copies those to the decoder, so background information is lost.
We remove the person from the inpainter's input by using the bodypart masks from our warping module.
Pixels not included in any of the projected bodypart masks become part of the background mask $M_\textit{BG}$.
The inpainting itself is performed using PartialConv layers~\cite{Liu18ECCV}.
The final result is a weighted combination (alpha blending) of the inpainted background $\hat I_\textit{BG}$ and the generated person $\hat I_\textit{FG}$ using $\hat M_\textit{FG}$ as the weights.

\PAR{Architectural details.} All our sub-networks except the background inpainter, but including the discriminator, are based on ResNet~\cite{He16CVPR,He16ECCV}.
We use GroupNorm~\cite{Wu18ECCV} instead of BatchNorm~\cite{Ioffe18ICML} due to its better performance with small batch sizes.
In $\etwo$ and $\dtwo$ we use bottleneck residual blocks to reduce computational cost.
Our 3D convolutional networks $\ethree$, $\dthree$ and $\epose$ do not use bottlenecks, as the number of features is already comparatively low.

\subsection{Training}
The perceptual loss $\loss{perc}$~\cite{Johnson16ECCV} compares generated and target image by passing both images through an ImageNet-pretrained VGG net~\cite{Simonyan15ICLR} and computing the $L_1$ loss on multiple feature maps.
The adversarial loss $\loss{adv}$ uses a discriminator net as in a classical GAN.
The discriminator gets the generated or ground truth image along with the input image and the 3D target heatmap.
We jointly optimize a weighted combination of these losses:
\vspace{-0.3mm}
\[\mathcal{L}(\theta) = \lambda_\textit{perc} \loss{perc}(\theta) + \lambda_\textit{adv} \loss{adv}(\theta)
\vspace{-0.3mm}
\]
We use data augmentation with rotation, scaling, translation, horizontal flip and color distortion.
We train with the Adam optimizer~\cite{Kingma15ICLR} for 150,000 steps with batch size 2 and learning rate $\alpha=2\cdot10^{-4}$.
We set $\lambda_{adv}=1$, $\lambda_{perc}=3$.

\section{Experiments}

\subsection{Datasets}

Commonly used in related work, the In-shop Clothes Retrieval Benchmark of the DeepFashion dataset (Fashion) \cite{Liu16CVPR} has almost 50,000 images and 8,000 sets of clothes.

The newer Impersonator dataset (iPER) \cite{Liu19ICCV} contains videos of 30 people and 103 clothing styles in total.
Two videos exist per clothing style, filmed from a static camera.
In one, the person turns around in an A-pose, the other shows arbitrary movements.

As these benchmark datasets do not supply 3D poses, we apply a 3D human pose estimation network based on~\cite{Sarandi20FG} to obtain the input and target poses $P_I$ and $P_T$.

\subsection{Evaluation Metrics}

Although generated image quality is somewhat subjective, several quantitative metrics have been used in related work to compare methods.
The structural similarity index (SSIM) \cite{Wang04TIP} compares patches of the generated image to patches of the ground truth according to luminance, contrast and structure.
While also used in some related work, we argue that the Inception score~\cite{Salimans16NIPS} is not suited for this task (we elaborate this in the supplementary).
We further use the learned perceptual image patch similarity (LPIPS)~\cite{Zhang2018CVPR}, which compares deep features between generated image and ground truth, similar to perceptual losses \cite{Johnson16ECCV}.

To evaluate high-level structure, we compare the response of a pretrained 3D pose estimator based on \cite{Sarandi20FG}, when applied to the generated and the true image.
We use the area under the PCK (percentage of correct keypoints) curve (AUC@150mm), a standard pose metric~\cite{Mehta17TDV}.

\subsection{Ablation Study}
\begin{table}
\footnotesize
\setlength\tabcolsep{1.6mm}
\begin{center}
\begin{tabularx}{\linewidth}{ccCccc} 
\toprule
3D warping & 3D target pose && SSIM $\uparrow$ & SSIM\textsubscript{fg} $\uparrow$ & Pose AUC $\uparrow$ \\ 
\midrule
--         & --         &&     0.872  &     0.566  &     0.698  \\
--         & \checkmark &&     0.875  &     0.578  &     0.749  \\
\checkmark & --         &&     0.877  &     0.607  &     0.749  \\
\checkmark & \checkmark && {\bf0.883} & {\bf0.626} & {\bf0.777} \\
\bottomrule
\end{tabularx}
\end{center}
\caption{
	Ablation on iPER.
	SSIM\textsubscript{fg} only evaluates the foreground.}
\label{tab:abl}
\end{table}

\begin{table}
	\footnotesize
	\setlength\tabcolsep{1.3mm}
	\begin{center}
		\begin{tabularx}{\linewidth}{Lrrrr}
			\toprule
			& \multicolumn{2}{c}{iPER} & \multicolumn{2}{c}{Fashion}\\
			& SSIM $\uparrow$ & LPIPS $\downarrow$ & SSIM $\uparrow$ & LPIPS $\downarrow$ \\
			\midrule
			PG2, Ma \etal \cite{Ma17NIPS}                      & 0.854 & 0.135 & 0.762 &    -- \\
			SHUP, Balakrishnan \etal \cite{Balakrishnan18CVPR} & 0.823 & 0.099 &    -- &    -- \\
			DSC, Siarohin \etal \cite{Siarohin18CVPR}          & 0.829 & 0.129 & 0.756 &    -- \\
			LWB, Liu \etal \cite{Liu19ICCV}                    & 0.840 & 0.087 &    -- &    -- \\
			SGW, Dong \etal \cite{Dong18NIPS}                  &    -- &    -- & 0.793 &    -- \\
			UPIS, Pumarola \etal \cite{Pumarola18CVPR}         &    -- &    -- & 0.747 &    -- \\
			VUNET, Esser \etal \cite{Esser18CVPR}              &    -- &    -- & 0.786 & 0.196 \\
			BodyROI7, Ma \etal \cite{Ma18CVPR}                 &    -- &    -- & 0.614 &    -- \\
			DPT, Neverova \etal \cite{Neverova18ECCV}          &    -- &    -- & 0.796 &    -- \\
			CTI, Grigorev \etal \cite{Grigorev19CVPR}          &    -- &    -- & 0.791 & {\bf0.169} \\
			Li \etal \cite{Li2019CVPR}                         &    -- &    -- & 0.778 &    -- \\
			\midrule
			Ours & {\bf0.883}\footnotemark & {\bf0.081}\footnotemark[1] & {\bf0.800} & 0.186 \\
			\bottomrule
		\end{tabularx}
	\end{center}
	\caption{
		Comparison to prior work.
		iPER scores taken from \cite{Liu19ICCV}.}
	\label{tab:prior}
\end{table}

\footnotetext{Update 29.05.20: with the evaluation method shared by \cite{Liu19ICCV} the results are {\bf0.863} for SSIM and {\bf0.077} for LPIPS.}

In contrast to prior work on person reposing, we propose to perform two different aspects in 3D: first, we use a 3D target pose and second, we perform 3D feature warping in the center of our model.
Architectural differences make it hard to directly compare our results to prior works, so we define ablation models to investigate these two aspects while keeping the exact same architecture otherwise.

To drop the depth information from the 3D target pose heatmaps, we project the pose to the image plane and replicate it to all depth layers.
Similarly, to perform warping in 2D, we project the body part masks to the image plane and copy them to all depths and apply 2D affine warpings to all depths independently.

The results on iPER (Tab.~\ref{tab:abl}) show that both of our 3D enhancements improve the scores compared to the 2D baseline and the results get even better when they are combined.
This is supported by the qualitative results (Tab.~\ref{tab:iperq}).
In the first row, the 2D pose models wrongly generate the right hand in front of the body, while the second row shows that a combination of both 3D aspects achieves the best results.

\subsection{Comparison to Prior Work}
Our model achieves state-of-the-art scores on both datasets (Tab.~\ref{tab:prior}).
Comparison to \cite{Liu19ICCV} on iPER (Tab.~\ref{tab:iperq}) shows that our model is able to transform the features of the left arm independently from the body features. In the upper row the hand correctly appears behind the body and the blue jacket in the lower row does not have a white stain as residue from the arm color.
On Fashion (Tab.~\ref{tab:fashq}), our model generates the overlapping arms of the right person better than the 2D feature warping approach of \cite{Siarohin18CVPR}.

Our architecture decreases the spatial size of the feature maps, which has the result that fine details are lost in some cases, which is also visible in the generated results. The buttons on the shirt in the first row of Tab.~\ref{tab:iperq} are missing in two ablation models and replaced by a zipper and shirt pockets in the other two.

\begin{table}
\footnotesize
\newcommand{\exampleipercross}[1]{\includegraphics[width=\linewidth,clip=true,trim=80 20 60 10]{imgs/examples/iper/cross/#1}}%
\begin{center}
\begin{tabularx}{\linewidth}{C@{\hspace{1pt}}C@{\hspace{3pt}}C@{\hspace{3pt}}C@{\hspace{1pt}}C@{\hspace{1pt}}C@{\hspace{3pt}}C}
\toprule  
\makecell{input\\image} & \makecell{target\\ pose} & \makecell{LWB \\ \cite{Liu19ICCV}} & 2D & \makecell{3D \\ pose} & \makecell{3D \\ warp} & \makecell{3D both \\(ours)} \\
\midrule
\exampleipercross{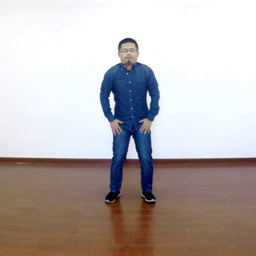} &
\exampleipercross{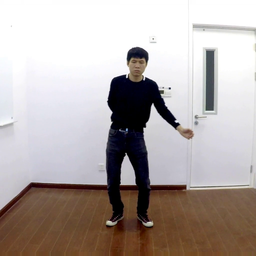} &
\exampleipercross{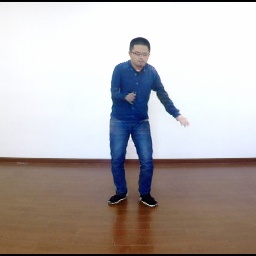} &
\exampleipercross{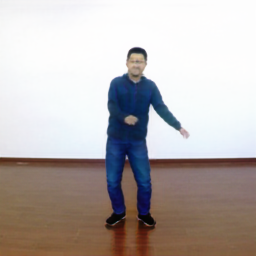} &
\exampleipercross{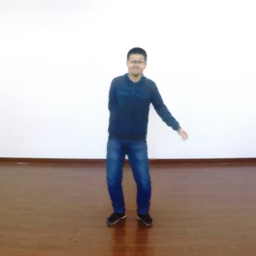} &
\exampleipercross{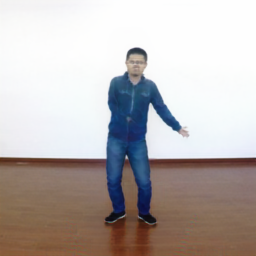} &
\exampleipercross{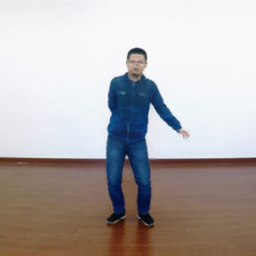}\\
\exampleipercross{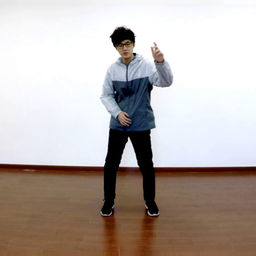} &
\exampleipercross{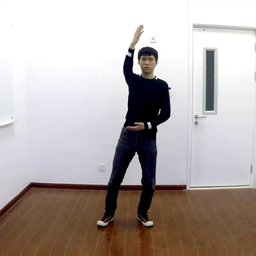} &
\exampleipercross{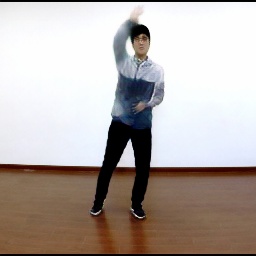} &
\exampleipercross{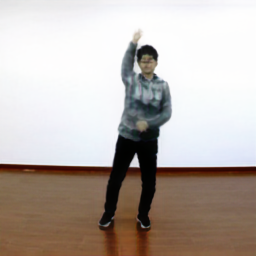} &
\exampleipercross{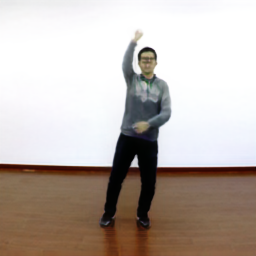} &
\exampleipercross{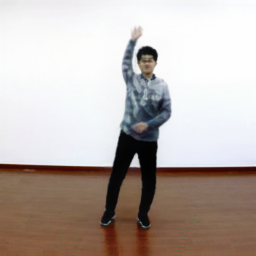} &
\exampleipercross{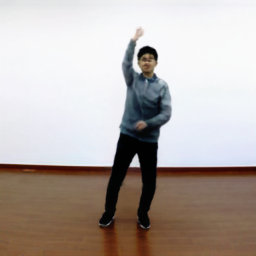}\\
\bottomrule
\end{tabularx}
\end{center}
\caption{Comparison to a mesh-based method and ablation models.}
\label{tab:iperq}
\end{table}

\begin{table}
\small
\newcommand{\examplefashtr}[1]{\includegraphics[width=\linewidth,clip=true,trim=50 0 50 0]{imgs/examples/fash/#1}}%
\newcommand{\examplefash}[1]{\includegraphics[width=\linewidth]{imgs/examples/fash/#1}}%
\begin{center}
\begin{tabularx}{\linewidth}{C@{\hspace{1pt}}C@{\hspace{1pt}}C@{\hspace{1pt}}C@{\hspace{3pt}}C@{\hspace{1pt}}C@{\hspace{1pt}}C@{\hspace{1pt}}C}
\toprule  
input image & target pose & DSC \cite{Siarohin18CVPR} & ours & input image & target pose & DSC \cite{Siarohin18CVPR} & ours \\
\midrule
\examplefashtr{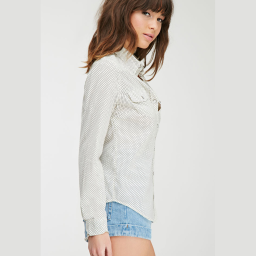} &
\examplefashtr{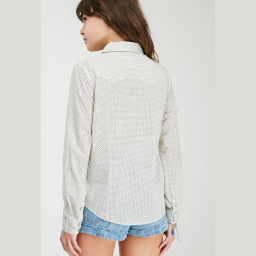} &
\examplefash{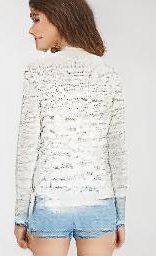} &
\examplefashtr{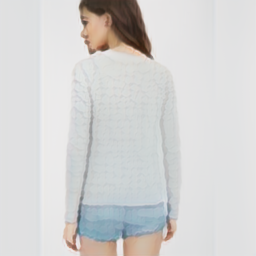} &
\examplefashtr{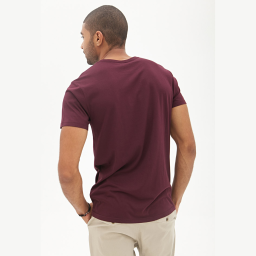} &
\examplefashtr{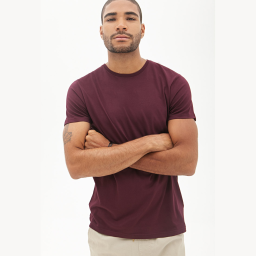} &
\examplefash{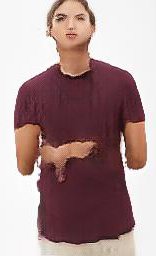} &
\examplefashtr{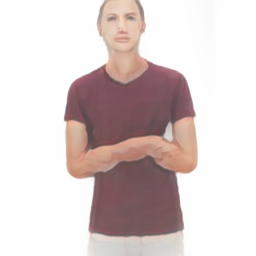}\\
\bottomrule
\end{tabularx}
\end{center}
\caption{Comparison with a 2D feature warping method. The target image is not used as input, only its pose.}
\label{tab:fashq}
\end{table}

\section{Conclusion}
We presented a novel architecture for person reposing, which relies on 3D warping of implicitly learned volumetric features.
Different from prior work, our approach is neither limited by approximating 3D motion with 2D transformations nor is an explicit 3D human mesh model required.

The ablation study and the comparison to related approaches showed that our method outperforms 2D warping methods by a significant margin.
This indicates that volumetric representations and 3D warping are a promising way to tackle reposing and we expect that more sophisticated neural rendering techniques could further improve results.

\ifcvprfinal
\PAR{Acknowledgments.}
This work was funded, in parts, by ERC Consolidator Grant project ``DeeViSe'' (ERC-CoG-2017-773161) and a Bosch Research Foundation grant. Some of the experiments were performed with computing resources granted by RWTH Aachen University under projects ``thes0618'' and ``rwth0479''.
\fi

{\small
\bibliographystyle{ieee_fullname}
\bibliography{abbrev_short,references}

\begin{thebibliography}{10}\itemsep=-1pt

\bibitem{Alp18CVPR}
R{\i}za Alp~G{\"u}ler, Natalia Neverova, and Iasonas Kokkinos.
\newblock Dense{P}ose: Dense human pose estimation in the wild.
\newblock In {\em CVPR}, 2018.

\bibitem{Balakrishnan18CVPR}
Guha Balakrishnan, Amy Zhao, Adrian~V Dalca, Fredo Durand, and John Guttag.
\newblock Synthesizing images of humans in unseen poses.
\newblock In {\em CVPR}, 2018.

\bibitem{Dong18NIPS}
Haoye Dong, Xiaodan Liang, Ke Gong, Hanjiang Lai, Jia Zhu, and Jian Yin.
\newblock Soft-gated warping-{GAN} for pose-guided person image synthesis.
\newblock In {\em NIPS}, 2018.

\bibitem{Esser18CVPR}
Patrick Esser, Ekaterina Sutter, and Bj{\"o}rn Ommer.
\newblock A variational {U}-net for conditional appearance and shape
  generation.
\newblock In {\em CVPR}, 2018.

\bibitem{Goodfellow14NIPS}
Ian Goodfellow et~al.
\newblock Generative adversarial nets.
\newblock In {\em NIPS}, 2014.

\bibitem{Grigorev19CVPR}
Artur Grigorev, Artem Sevastopolsky, Alexander Vakhitov, and Victor Lempitsky.
\newblock Coordinate-based texture inpainting for pose-guided human image
  generation.
\newblock In {\em CVPR}, 2019.

\bibitem{He16CVPR}
Kaiming He, Xiangyu Zhang, Shaoqing Ren, and Jian Sun.
\newblock Deep residual learning for image recognition.
\newblock In {\em CVPR}, 2016.

\bibitem{He16ECCV}
Kaiming He, Xiangyu Zhang, Shaoqing Ren, and Jian Sun.
\newblock Identity mappings in deep residual networks.
\newblock In {\em ECCV}, 2016.

\bibitem{Horiuchi19MVA}
Yusuke Horiuchi, Satoshi Iizuka, Edgar Simo-Serra, and Hiroshi Ishikawa.
\newblock Spectral normalization and relativistic adversarial training for
  conditional pose generation with self-attention.
\newblock In {\em MVA}, 2019.

\bibitem{Lakhal18ECCV}
Mohamed Ilyes~Lakhal, Oswald Lanz, and Andrea Cavallaro.
\newblock Pose guided human image synthesis by view disentanglement and
  enhanced weighting loss.
\newblock In {\em ECCV}, 2018.

\bibitem{Ioffe18ICML}
Sergey Ioffe and Christian Szegedy.
\newblock Batch normalization: Accelerating deep network training by reducing
  internal covariate shift.
\newblock In {\em ICML}, 2015.

\bibitem{Isola17CVPR}
Phillip Isola, Jun-Yan Zhu, Tinghui Zhou, and Alexei~A Efros.
\newblock Image-to-image translation with conditional adversarial networks.
\newblock In {\em CVPR}, 2017.

\bibitem{Johnson16ECCV}
Justin Johnson, Alexandre Alahi, and Li Fei-Fei.
\newblock Perceptual losses for real-time style transfer and super-resolution.
\newblock In {\em ECCV}, 2016.

\bibitem{Kingma15ICLR}
Diederik~P Kingma and Jimmy Ba.
\newblock Adam: A method for stochastic optimization.
\newblock In {\em ICLR}, 2015.

\bibitem{Li2019CVPR}
Yining Li, Chen Huang, and Chen~Change Loy.
\newblock Dense intrinsic appearance flow for human pose transfer.
\newblock In {\em CVPR}, 2019.

\bibitem{Liu18ECCV}
Guilin Liu, Fitsum~A Reda, Kevin~J Shih, Ting-Chun Wang, Andrew Tao, and Bryan
  Catanzaro.
\newblock Image inpainting for irregular holes using partial convolutions.
\newblock In {\em ECCV}, 2018.

\bibitem{Liu19ICCV}
Wen Liu, Zhixin Piao, Jie Min, Wenhan Luo, Lin Ma, and Shenghua Gao.
\newblock Liquid warping {GAN}: A unified framework for human motion imitation,
  appearance transfer and novel view synthesis.
\newblock In {\em ICCV}, 2019.

\bibitem{Liu16CVPR}
Ziwei Liu, Ping Luo, Shi Qiu, Xiaogang Wang, and Xiaoou Tang.
\newblock Deep{F}ashion: Powering robust clothes recognition and retrieval with
  rich annotations.
\newblock In {\em CVPR}, 2016.

\bibitem{Luvizon18CVPR}
Diogo~C. Luvizon, David Picard, and Hedi Tabia.
\newblock 2{D}/3{D} pose estimation and action recognition using multitask deep
  learning.
\newblock In {\em CVPR}, 2018.

\bibitem{Ma17NIPS}
Liqian Ma, Xu Jia, Qianru Sun, Bernt Schiele, Tinne Tuytelaars, and Luc
  Van~Gool.
\newblock Pose guided person image generation.
\newblock In {\em NIPS}, 2017.

\bibitem{Ma18CVPR}
Liqian Ma, Qianru Sun, Stamatios Georgoulis, Luc Van~Gool, Bernt Schiele, and
  Mario Fritz.
\newblock Disentangled person image generation.
\newblock In {\em CVPR}, 2018.

\bibitem{Mehta17TDV}
Dushyant Mehta, Helge Rhodin, Dan Casas, Pascal Fua, Oleksandr Sotnychenko,
  Weipeng Xu, and Christian Theobalt.
\newblock Monocular 3{D} human pose estimation in the wild using improved {CNN}
  supervision.
\newblock In {\em 3DV}, 2017.

\bibitem{Neverova18ECCV}
Natalia Neverova, Riza Alp~Guler, and Iasonas Kokkinos.
\newblock Dense pose transfer.
\newblock In {\em ECCV}, 2018.

\bibitem{Nguyen19ICCVW}
Thu Nguyen-Phuoc, Chuan Li, Lucas Theis, Christian Richardt, and Yong-Liang
  Yang.
\newblock Holo{GAN}: Unsupervised learning of 3d representations from natural
  images.
\newblock 2019.

\bibitem{Pavlakos17CVPR}
Georgios Pavlakos, Xiaowei Zhou, Konstantinos~G Derpanis, and Kostas
  Daniilidis.
\newblock Coarse-to-fine volumetric prediction for single-image 3{D} human
  pose.
\newblock In {\em CVPR}, 2017.

\bibitem{Pumarola18CVPR}
Albert Pumarola, Antonio Agudo, Alberto Sanfeliu, and Francesc Moreno-Noguer.
\newblock Unsupervised person image synthesis in arbitrary poses.
\newblock In {\em CVPR}, 2018.

\bibitem{Rhodin18ECCV}
Helge Rhodin, Mathieu Salzmann, and Pascal Fua.
\newblock Unsupervised geometry-aware representation for 3{D} human pose
  estimation.
\newblock In {\em ECCV}, 2018.

\bibitem{Ronneberger15MICCAI}
Olaf Ronneberger, Philipp Fischer, and Thomas Brox.
\newblock U-net: Convolutional networks for biomedical image segmentation.
\newblock In {\em MICCAI}, 2015.

\bibitem{Salimans16NIPS}
Tim Salimans, Ian Goodfellow, Wojciech Zaremba, Vicki Cheung, Alec Radford, and
  Xi Chen.
\newblock Improved techniques for training {GAN}s.
\newblock In {\em NIPS}, 2016.

\bibitem{Sarandi20FG}
Istv\'an S\'ar\'andi, Timm Linder, Kai~O. Arras, and Bastian Leibe.
\newblock Metric-scale truncation-robust heatmaps for 3{D} human pose
  estimation.
\newblock In {\em Int. Conf. on Automatic Face and Gesture Recognition}, 2020.

\bibitem{Si18CVPR}
Chenyang Si, Wei Wang, Liang Wang, and Tieniu Tan.
\newblock Multistage adversarial losses for pose-based human image synthesis.
\newblock In {\em CVPR}, 2018.

\bibitem{Siarohin18CVPR}
Aliaksandr Siarohin, Enver Sangineto, St{\'e}phane Lathuili{\`e}re, and Nicu
  Sebe.
\newblock Deformable {GAN}s for pose-based human image generation.
\newblock In {\em CVPR}, 2018.

\bibitem{Simonyan15ICLR}
Karen Simonyan and Andrew Zisserman.
\newblock Very deep convolutional networks for large-scale image recognition.
\newblock In {\em ICLR}, 2015.

\bibitem{Sitzmann19CVPR}
Vincent Sitzmann, Justus Thies, Felix Heide, Matthias Nie{\ss}ner, Gordon
  Wetzstein, and Michael Zollhofer.
\newblock {D}eep{V}oxels: Learning persistent 3d feature embeddings.
\newblock In {\em CVPR}, 2019.

\bibitem{Sun18ECCV}
Xiao Sun, Bin Xiao, Shuang Liang, and Yichen Wei.
\newblock Integral human pose regression.
\newblock In {\em ECCV}, 2018.

\bibitem{Wang04TIP}
Zhou Wang, Alan~C Bovik, Hamid~R Sheikh, Eero~P Simoncelli, et~al.
\newblock Image quality assessment: from error visibility to structural
  similarity.
\newblock {\em Trans. Image Proc.}, 2004.

\bibitem{Wu18ECCV}
Yuxin Wu and Kaiming He.
\newblock Group normalization.
\newblock In {\em ECCV}, 2018.

\bibitem{Zanfir18CVPR}
Mihai Zanfir, Alin-Ionut Popa, Andrei Zanfir, and Cristian Sminchisescu.
\newblock Human appearance transfer.
\newblock In {\em CVPR}, 2018.

\bibitem{Zhang2018CVPR}
Richard Zhang, Phillip Isola, Alexei~A Efros, Eli Shechtman, and Oliver Wang.
\newblock The unreasonable effectiveness of deep features as a perceptual
  metric.
\newblock In {\em CVPR}, 2018.

\bibitem{Zhu19CVPR}
Zhen Zhu, Tengteng Huang, Baoguang Shi, Miao Yu, Bofei Wang, and Xiang Bai.
\newblock Progressive pose attention transfer for person image generation.
\newblock In {\em CVPR}, 2019.

\end{thebibliography}


\begin{thebibliography}{1}\itemsep=-1pt

\bibitem{Liu19ICCV}
Wen Liu, Zhixin Piao, Jie Min, Wenhan Luo, Lin Ma, and Shenghua Gao.
\newblock Liquid warping {GAN}: A unified framework for human motion imitation,
  appearance transfer and novel view synthesis.
\newblock In {\em ICCV}, 2019.

\bibitem{Russakovsky15IJCV}
Olga Russakovsky et~al.
\newblock Image{N}et large scale visual recognition challenge.
\newblock {\em IJCV}, 2015.

\bibitem{Salimans16NIPS}
Tim Salimans, Ian Goodfellow, Wojciech Zaremba, Vicki Cheung, Alec Radford, and
  Xi Chen.
\newblock Improved techniques for training {GAN}s.
\newblock In {\em NIPS}, 2016.

\bibitem{Siarohin18CVPR}
Aliaksandr Siarohin, Enver Sangineto, St{\'e}phane Lathuili{\`e}re, and Nicu
  Sebe.
\newblock Deformable {GAN}s for pose-based human image generation.
\newblock In {\em CVPR}, 2018.

\bibitem{Szegedy16CVPR}
Christian Szegedy, Vincent Vanhoucke, Sergey Ioffe, Jon Shlens, and Zbigniew
  Wojna.
\newblock Rethinking the inception architecture for computer vision.
\newblock In {\em CVPR}, 2016.

\end{thebibliography}
}

\end{document}


\title{Reposing Humans by Warping 3D Features\\
(Supplementary Material)}

\author{Markus Knoche \qquad Istv\'an S\'ar\'andi \qquad Bastian Leibe\\
RWTH Aachen University, Germany\\
{\tt\small \{knoche,sarandi,leibe\}@vision.rwth-aachen.de}
}

\maketitle
\thispagestyle{empty}

\begin{appendix}

\section{Inception Score's Unsuitability in Reposing}

Many related works use the Inception score (IS) \cite{Salimans16NIPS}, as a metric for person reposing.
IS was proposed to evaluate unconditioned GANs, \ie, GANs which are supposed to generate a diverse dataset like ImageNet \cite{Russakovsky15IJCV} based on random inputs.
Two aspects are combined: realism of single generated images and variability of a large set of generated images.
Generated images are passed through the Inception network \cite{Szegedy16CVPR}, a single realistic image $x$ should be confidently assigned to a single class, so the assigned label distribution $p(y|x)$ has a single high activation.
In contrast, multiple generated images should belong to different classes, thus $p(y)$ is rather uniform.
IS compares these distributions using the Kullback-Leibler divergence, which means that the score is high, if the distributions are dissimilar.
For human reposing only a single output class exists, such that for a perfect generator both $p(y)$ and $p(y|x)$ are the same, because the Inception network always assigns the label ``human''.
This issue invalidates the Inception Score as a metric for person reposing.

\section{Evaluation Protocol on iPER}

Liu \etal~\cite{Liu19ICCV}, who published the iPER dataset, perform evaluation by selecting three frames per person and then generating the full video based on each of these frames.
The results are then compared to the original videos using the quantitative metrics.

As the authors have not published their frame selection procedure, replicating their exact evaluation protocol is currently not possible.
We therefore use the following selection procedure: we first uniformly sample a random clothing layout from the test set, then randomly select two frames from this person.
The network then generates the second frame based from the first one.
This process is repeated 10,000 times, the mean scores are reported.

Update 29.05.2020: The evaluation procedure by Liu \etal~\cite{Liu19ICCV} has become available recently. We have re-evaluated our final model according to their scheme and included the results as a footnote.

\section{Additional Qualitative Results}

We compare our model to \cite[supplementary Figure 9]{Siarohin18CVPR} in Table \ref{tab:fasha}.

Table \ref{tab:ipera} shows generated images of our model compared to the ablation models and to the results of \cite[Figure 7]{Liu19ICCV}.

{\small
\bibliographystyle{ieee_fullname}
\bibliography{abbrev_short,references}
}

\begin{table}
\footnotesize
\newcommand{\examplefashtr}[1]{\includegraphics[width=\linewidth,clip=true,trim=50 0 50 0]{imgs/examples/fash/#1.jpg}}%
\newcommand{\examplefash}[1]{\includegraphics[width=\linewidth]{imgs/examples/fash/#1.jpg}}%
\begin{center}
\begin{tabularx}{\linewidth}{C@{\hspace{1pt}}C@{\hspace{1pt}}C@{\hspace{1pt}}C}
\toprule  
input image & target pose & DSC \cite{Siarohin18CVPR} & ours\\
\midrule
\examplefashtr{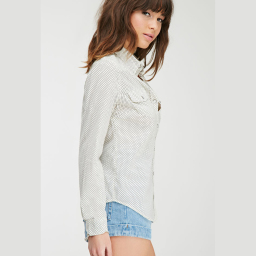} &
\examplefashtr{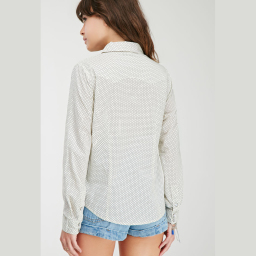} &
\examplefash{siar1} &
\examplefashtr{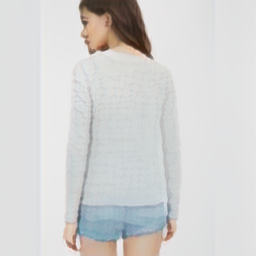}\\
\examplefashtr{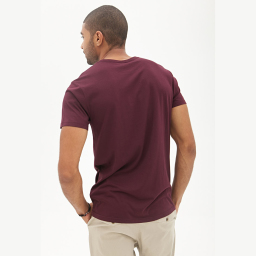} &
\examplefashtr{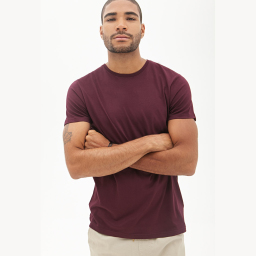} &
\examplefash{siar2} &
\examplefashtr{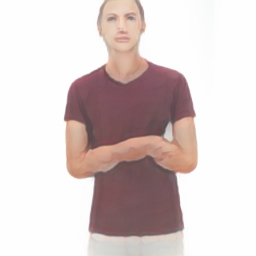}\\
\examplefashtr{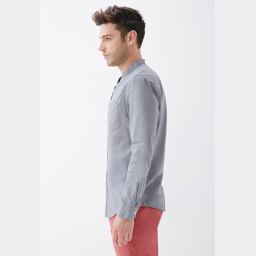} &
\examplefashtr{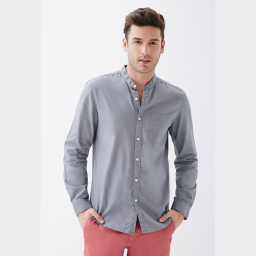} &
\examplefash{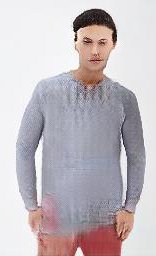} &
\examplefashtr{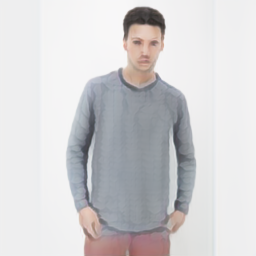}\\
\examplefashtr{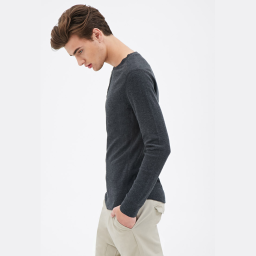} &
\examplefashtr{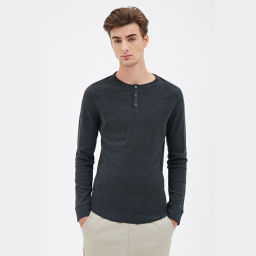} &
\examplefash{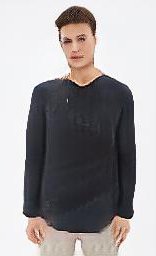} &
\examplefashtr{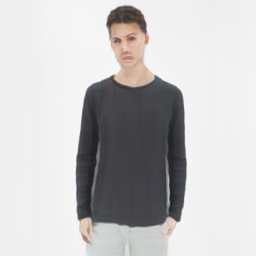}\\
\examplefashtr{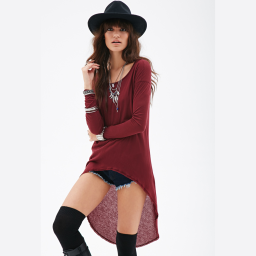} &
\examplefashtr{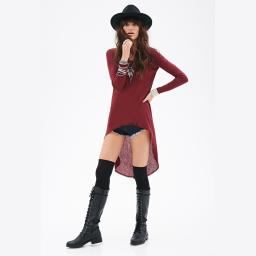} &
\examplefash{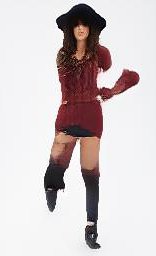} &
\examplefashtr{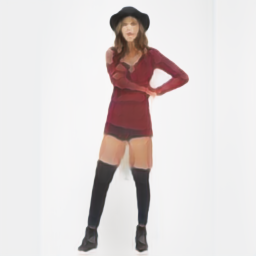}\\
\examplefashtr{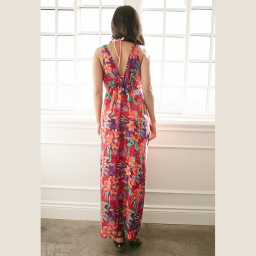} &
\examplefashtr{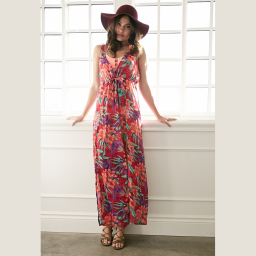} &
\examplefash{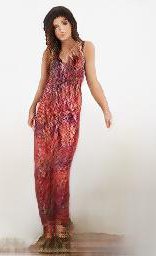} &
\examplefashtr{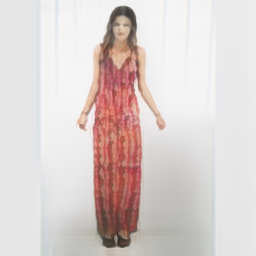}\\
\bottomrule
\end{tabularx}
\end{center}
\caption{Qualitative comparison with a 2D feature warping method. The target image is not used as input, only its pose.}
\label{tab:fasha}
\end{table}

\begin{table}
\footnotesize
\newcommand{\exampleipercross}[1]{\includegraphics[width=\linewidth,clip=true,trim=20 20 20 10]{imgs/examples/iper/cross/#1.jpg}}%
\begin{center}
\begin{tabularx}{\linewidth}{C@{\hspace{1pt}}C@{\hspace{3pt}}C@{\hspace{3pt}}C@{\hspace{1pt}}C@{\hspace{1pt}}C@{\hspace{3pt}}C}
\toprule  
input image & target pose & LWB \cite{Liu19ICCV} & 2D & 3D pose & 3D warp & 3D both (ours) \\
\midrule
\exampleipercross{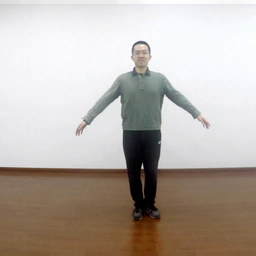} &
\exampleipercross{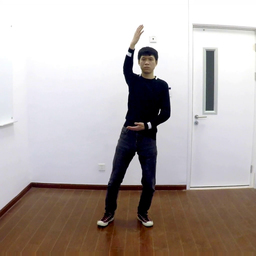} &
\exampleipercross{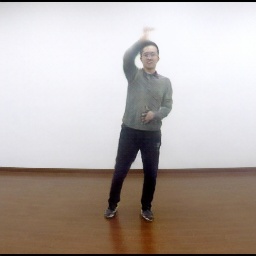} &
\exampleipercross{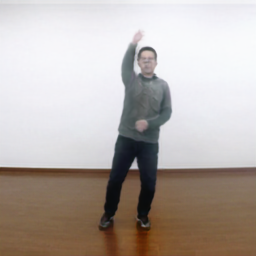} &
\exampleipercross{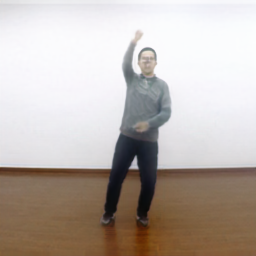} &
\exampleipercross{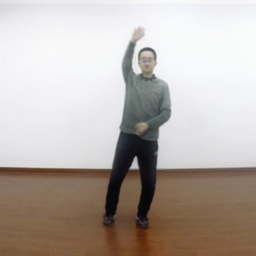} &
\exampleipercross{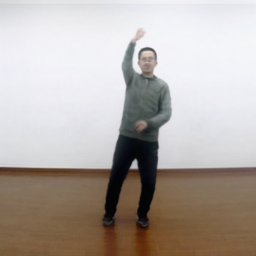}\\
&
\exampleipercross{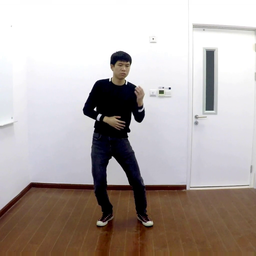} &
\exampleipercross{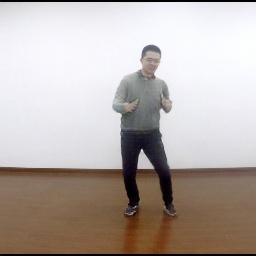} &
\exampleipercross{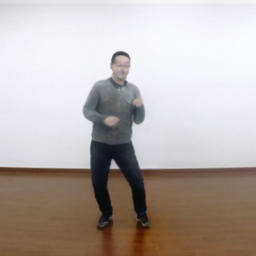} &
\exampleipercross{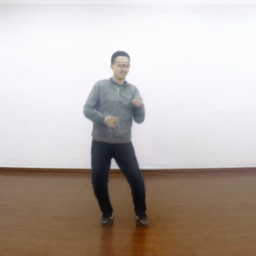} &
\exampleipercross{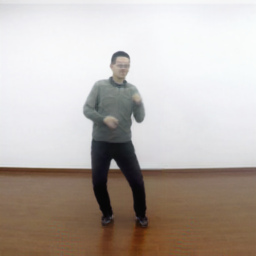} &
\exampleipercross{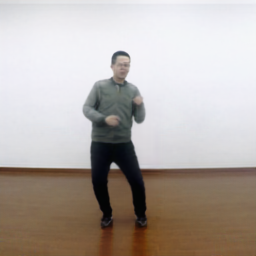}\\
&
\exampleipercross{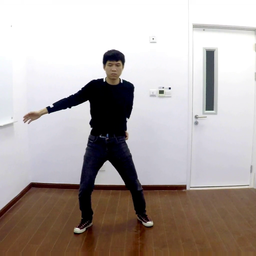} &
\exampleipercross{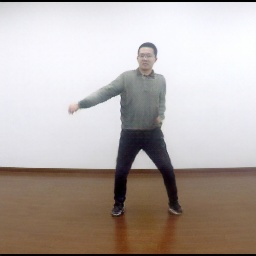} &
\exampleipercross{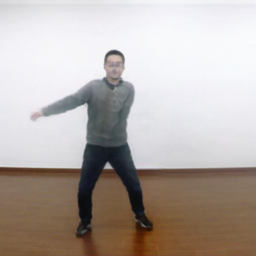} &
\exampleipercross{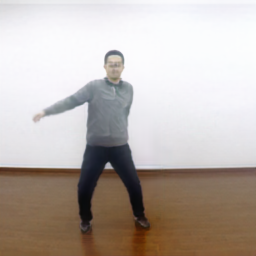} &
\exampleipercross{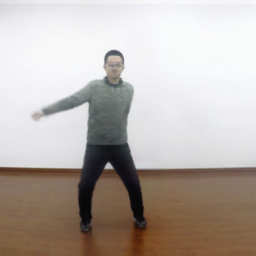} &
\exampleipercross{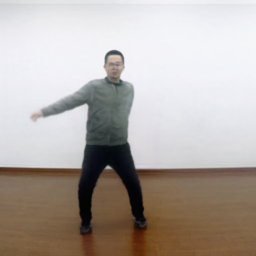}\\
&
\exampleipercross{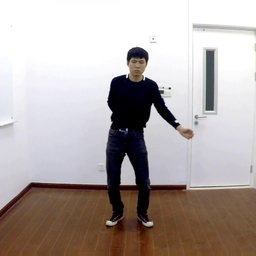} &
\exampleipercross{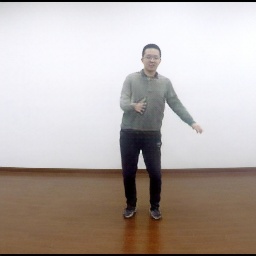} &
\exampleipercross{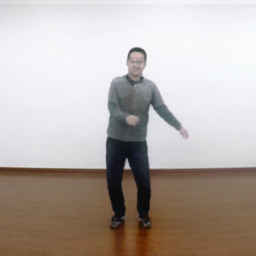} &
\exampleipercross{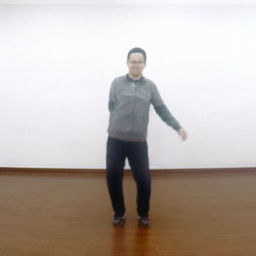} &
\exampleipercross{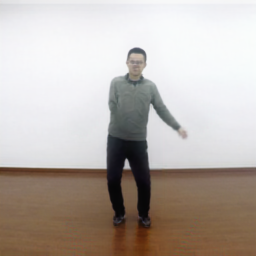} &
\exampleipercross{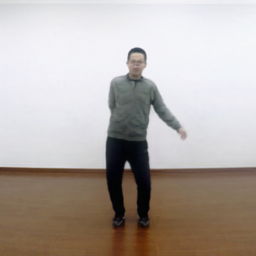}\\
&
\exampleipercross{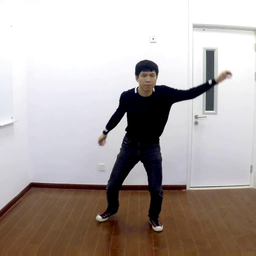} &
\exampleipercross{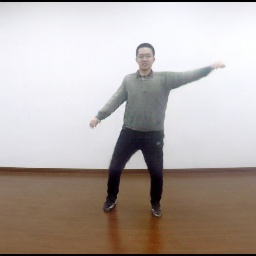} &
\exampleipercross{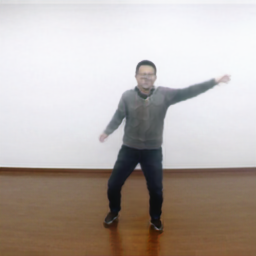} &
\exampleipercross{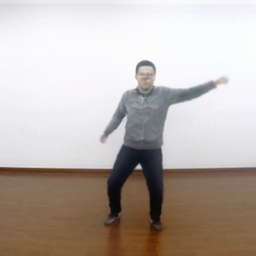} &
\exampleipercross{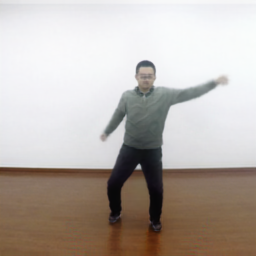} &
\exampleipercross{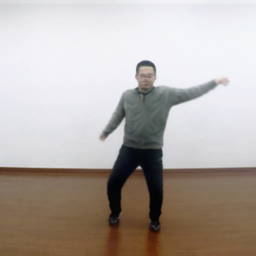}\\
\exampleipercross{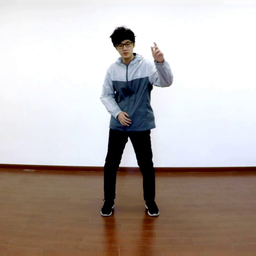} &
\exampleipercross{to1} &
\exampleipercross{iper21} &
\exampleipercross{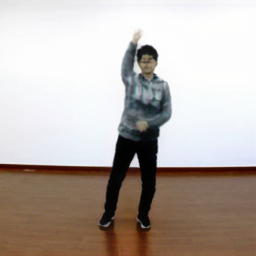} &
\exampleipercross{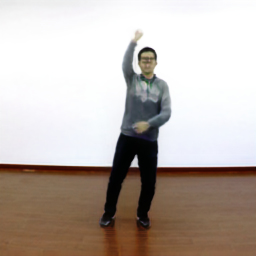} &
\exampleipercross{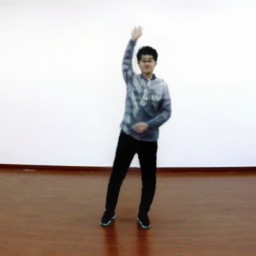} &
\exampleipercross{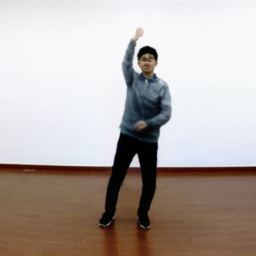}\\
&
\exampleipercross{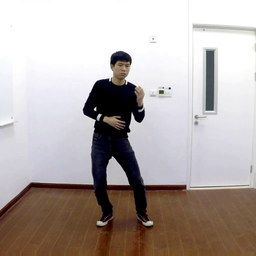} &
\exampleipercross{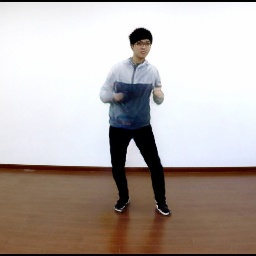} &
\exampleipercross{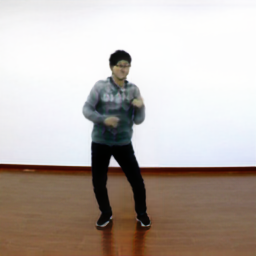} &
\exampleipercross{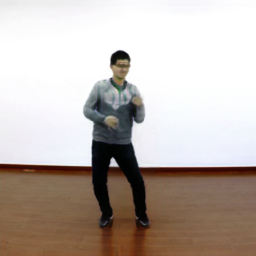} &
\exampleipercross{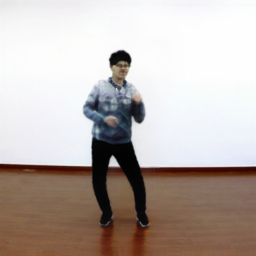} &
\exampleipercross{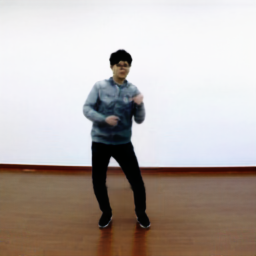}\\
&
\exampleipercross{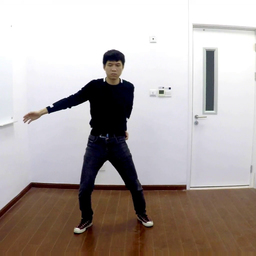} &
\exampleipercross{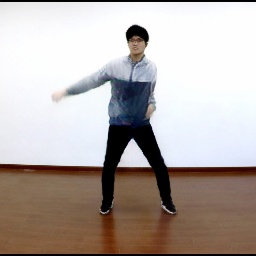} &
\exampleipercross{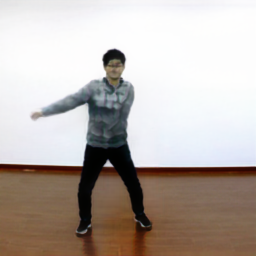} &
\exampleipercross{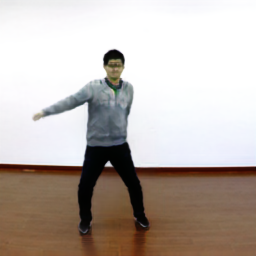} &
\exampleipercross{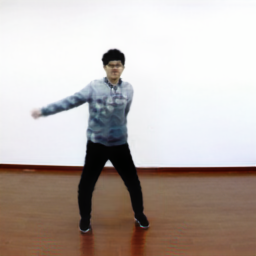} &
\exampleipercross{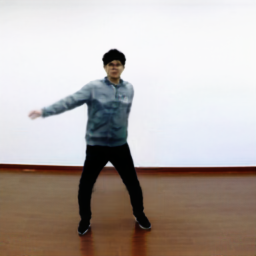}\\
&
\exampleipercross{to4} &
\exampleipercross{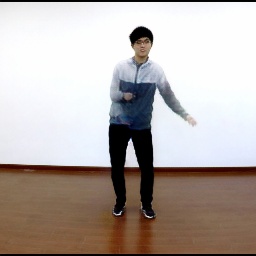} &
\exampleipercross{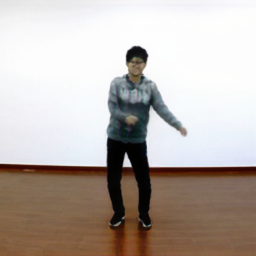} &
\exampleipercross{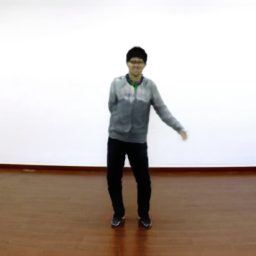} &
\exampleipercross{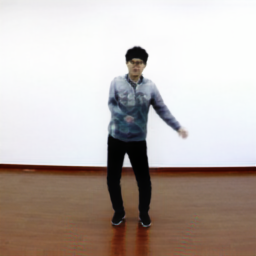} &
\exampleipercross{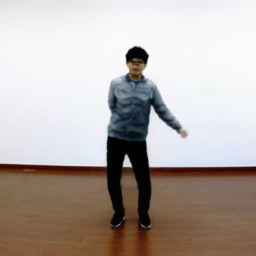}\\
&
\exampleipercross{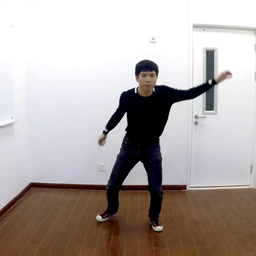} &
\exampleipercross{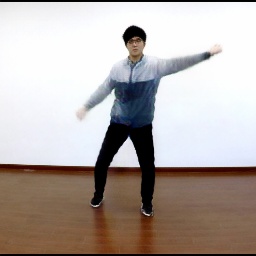} &
\exampleipercross{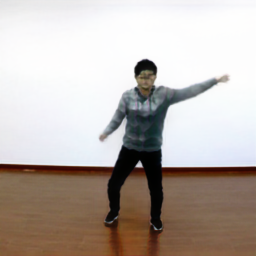} &
\exampleipercross{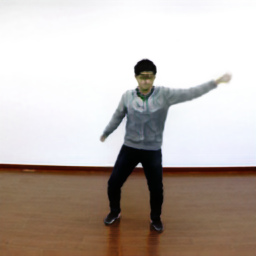} &
\exampleipercross{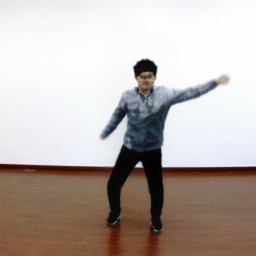} &
\exampleipercross{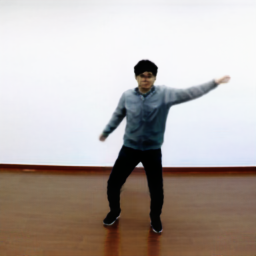}\\
\exampleipercross{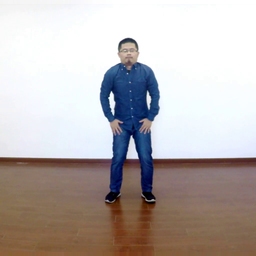} &
\exampleipercross{to1} &
\exampleipercross{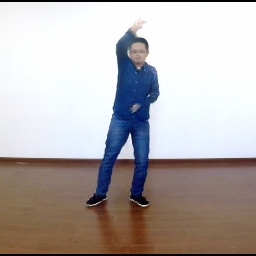} &
\exampleipercross{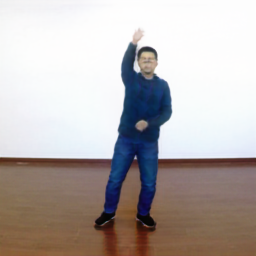} &
\exampleipercross{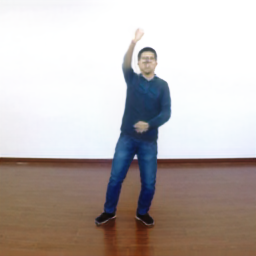} &
\exampleipercross{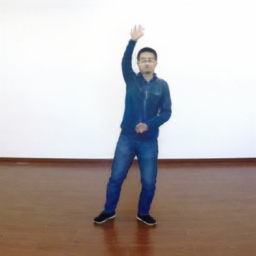} &
\exampleipercross{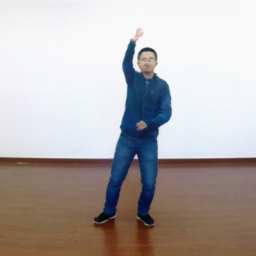}\\
&
\exampleipercross{to2} &
\exampleipercross{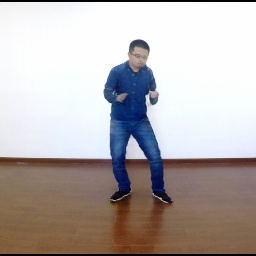} &
\exampleipercross{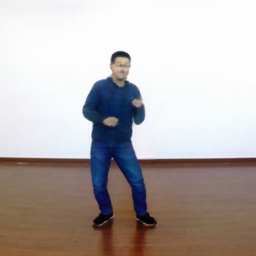} &
\exampleipercross{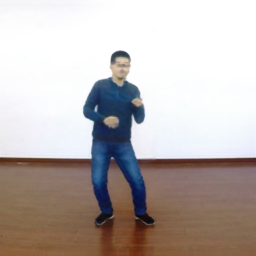} &
\exampleipercross{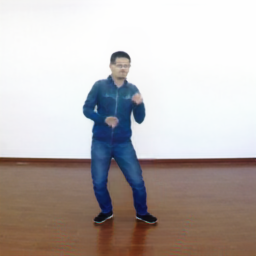} &
\exampleipercross{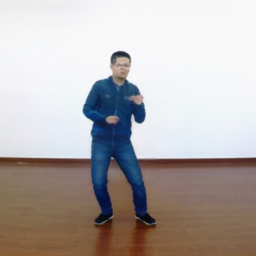}\\
&
\exampleipercross{to3} &
\exampleipercross{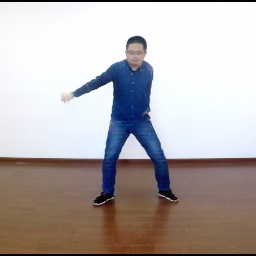} &
\exampleipercross{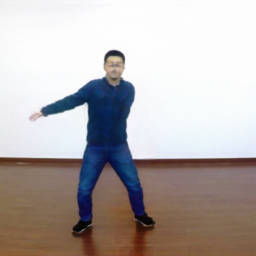} &
\exampleipercross{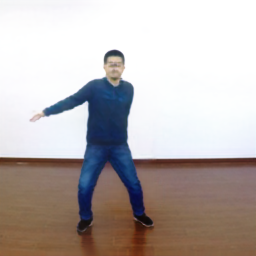} &
\exampleipercross{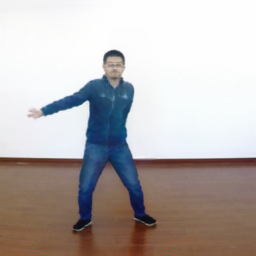} &
\exampleipercross{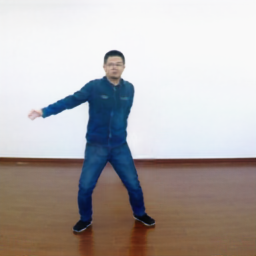}\\
&
\exampleipercross{to4} &
\exampleipercross{iper34} &
\exampleipercross{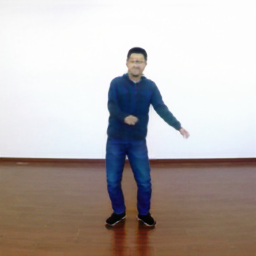} &
\exampleipercross{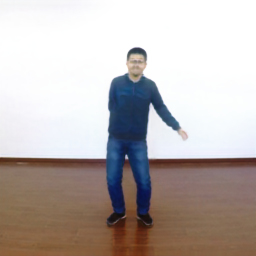} &
\exampleipercross{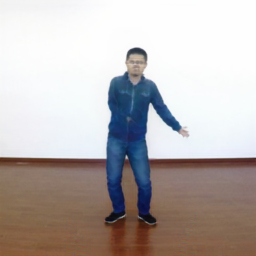} &
\exampleipercross{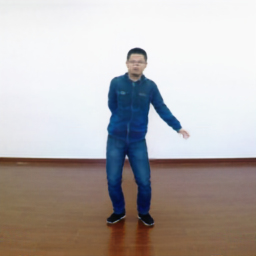}\\
&
\exampleipercross{to5} &
\exampleipercross{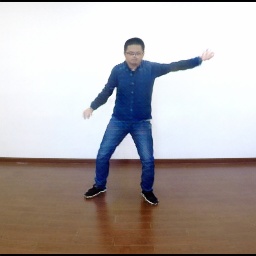} &
\exampleipercross{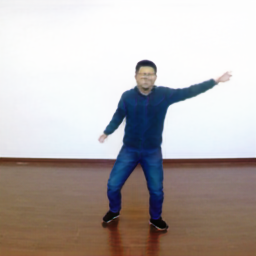} &
\exampleipercross{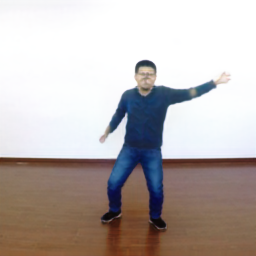} &
\exampleipercross{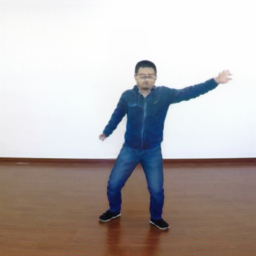} &
\exampleipercross{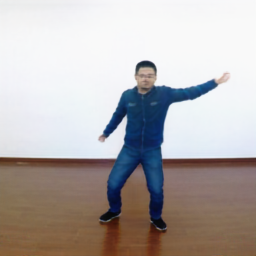}\\
\bottomrule
\end{tabularx}
\end{center}
\caption{Qualitative comparison to a mesh-based approach and to our ablation models.}
\label{tab:ipera}
\end{table}

\end{appendix}